\pdfoutput=1
\documentclass[10pt,twocolumn,letterpaper]{article}

\usepackage{iccv}
\usepackage{times}
\usepackage{epsfig}
\usepackage{graphicx}
\usepackage{amsmath}
\usepackage{mathrsfs}
\usepackage{amssymb}
\usepackage{tabulary}
\usepackage{color}
\usepackage{url}
\usepackage{multirow}
\usepackage{bm}
\usepackage{makecell}
\usepackage{bbding}
\usepackage{threeparttable}
\usepackage[pagebackref,breaklinks,colorlinks]{hyperref}

\iccvfinalcopy % *** Uncomment this line for the final submission

\def\iccvPaperID{7281} % *** Enter the ICCV Paper ID here

\ificcvfinal\fi

\usepackage[capitalize]{cleveref}
\crefname{section}{Sec.}{Secs.}
\Crefname{section}{Section}{Sections}
\Crefname{table}{Table}{Tables}
\crefname{table}{Tab.}{Tabs.}

\definecolor{turquoise}{cmyk}{0.65,0,0.1,0.1}
\definecolor{purple}{rgb}{0.65,0,0.65}
\definecolor{dark_green}{rgb}{0, 0.5, 0}
\definecolor{orange}{rgb}{0.8, 0.6, 0.2}
\definecolor{red}{rgb}{0.8, 0.2, 0.2}
\definecolor{brown}{rgb}{0.5, 0.16, 0.16}

\def\iccvPaperID{7281} % *** Enter the ICCV Paper ID here

\ificcvfinal\fi

\begin{document}

%%%%%%%%% TITLE
\title{CVSformer: Cross-View Synthesis Transformer for Semantic Scene Completion}

\author{
Haotian Dong\textsuperscript{1}\thanks{Co-first authors. The names are listed in alphabetical order.}, \enskip Enhui Ma\textsuperscript{1}\footnotemark[1], \enskip Lubo Wang\textsuperscript{1}, \enskip Miaohui Wang\textsuperscript{2}, \enskip Wuyuan Xie\textsuperscript{2},\\ Qing Guo\textsuperscript{3}, \enskip Ping Li \textsuperscript{4}, \enskip Lingyu Liang\textsuperscript{5}\thanks{Co-corresponding authors.
}, \enskip Kairui Yang\textsuperscript{6},  \enskip Di Lin \textsuperscript{1}\footnotemark[2]
\\ \\
    \textsuperscript{1}Tianjin University,
     \enskip \textsuperscript{2} Shenzhen University, \enskip
    \textsuperscript{3}A*STAR,\\
    \textsuperscript{4}
    The Hong Kong Polytechnic University,\\
    \textsuperscript{5}
    South China University of Technology,\\
    \textsuperscript{6}
    Alibaba Damo Academy
}
\maketitle
\ificcvfinal\thispagestyle{empty}\fi
%\thispagestyle{empty}

%%%%%%%%% ABSTRACT
%\begin{abstract}
%
%\end{abstract}

%%%%%%%%% BODY TEXT
\begin{abstract}

    Semantic scene completion (SSC) requires an accurate understanding of the geometric and semantic relationships between the objects in the 3D scene for reasoning the occluded objects. The popular SSC methods voxelize the 3D objects, allowing the deep 3D convolutional network (3D CNN) to learn the object relationships from the complex scenes. However, the current networks lack the controllable kernels to model the object relationship across multiple views, where appropriate views provide the relevant information for suggesting the existence of the occluded objects. In this paper, we propose \textbf{Cross-View Synthesis Transformer} (CVSformer), which consists of \textbf{Multi-View Feature Synthesis} and \textbf{Cross-View Transformer} for learning cross-view object relationships. In the multi-view feature synthesis, we use a set of 3D convolutional kernels rotated differently to compute the multi-view features for each voxel. In the cross-view transformer, we employ the cross-view fusion to comprehensively learn the cross-view relationships, which form useful information for enhancing the features of individual views. We use the enhanced features to predict the geometric occupancies and semantic labels of all voxels. We evaluate CVSformer on public datasets, where CVSformer yields state-of-the-art results.

    %Single-image rain removal requires the accurate separation between the pixels of the rain streaks and object textures. But the confusing appearances of rains and objects lead to the misunderstanding of pixels, thus remaining the rain streaks or missing the object textures in the result. In this paper, we propose \textbf{SEIDNet} equipped with the pixel-wise \textbf{Status Estimation} and \textbf{Information Decoupling} for deraining. In the status estimation, we embed the pixel-wise statuses into the status space, where each status indicates a pixel of the rain or object. In the information decoupling, we respect the statuses of different pixels and decouple the information of rains and objects from the image. Based on the decoupled information, we learn the independent convolutional kernels for removing the rains and preserving the objects. We embed these kernels into the kernel space.
    %
    %For the deraining task, we use pixel-wise information as the condition and multiple statuses from the status space. These statuses capture the confusing information of the pixel. They play as the condition for sampling from the kernel space, yielding multiple kernels to remove the rain and preserve the object information of the pixel. We evaluate SEIDNet on the public datasets, achieving state-of-the-art performances of image deraining.

    \end{abstract} 

\section{Introduction}

%Many artificial intelligence (AI) systems (e.g., autonomous robot and video surveillance system) rely on the images captured by the cameras, for recognizing the objects in the outdoor scenes. The outdoor-scene object recognition is heavily influenced by the weather conditions. Especially, the images taken on the rainy days contain the rain streaks that are irrelevant to the objects. The rain streaks distract the object recognition, thus giving rise to the unexpected behaviours of the AI systems.

The recent progress on artificial intelligence is mainly driven by the advanced machine power of recognizing 3D objects. It significantly benefits various downstream applications, such as autonomous driving and video surveillance. An essential problem of 3D object recognition is letting the machine accurately recognize the occluded objects in complex 3D scenes. This problem leads to the emergence of research on semantic scene completion (SSC).

%The core of SSC lies in understanding the geometric and semantic relationships between 3D objects.

The latest success of SSC is primarily attributed to deep neural networks, which are good at learning the geometric and semantic representations of 3D objects. To facilitate fast representation learning in the large-scale 3D scene, the popular SSC methods~\cite{3dsketch,3dssc,edgenet,two,nyucad,palnet,sscnet,sgc,ccpnet,grfnet,li2020attention,roldao2020lmscnet,zhang2018semantic,wang2018adversarial,cao2022monoscene,dourado2022data,miao2023occdepth, liang2021sscnav,li2023voxformer,li2021imenet, dourado2020semantic} employ 3D CNN to learn representations from the voxelized objects. To alleviate the limitation of the regular 3D convolutional kernels that fix the range of capturing the object relationship, the variant 3D CNNs involve spatial pyramid~\cite{ccpnet} and deformation~\cite{d3d} to diversify the kernel shapes, which attend to the object relationships in different ranges. Yet, the pyramidal kernels lack the flexibility to exclude the irrelevant voxels; the deformable kernels are sensitive to the object shapes, usually capturing the relationship between voxels from a mono-view. These kernels work without the controllable view like the camera. They are also less effective for modeling the object relationship across multiple perspectives, like multiple cameras, which enable the change of view directions to offer the information of relevant objects for recognizing the occluded things.

%Though the variant kernels with various shapes to process the different object voxels, only a single shape captures the relationship between each voxel and its neighbors from a mono-view.

%\begin{figure*}[t!]
%\centering
%\includegraphics[width=\linewidth]{Images/overview/overview_Final.pdf}
%\vspace{-0.2in}
%\caption{The overall network of the proposed method. Given a pair of single-view RGB-D images, the segmentation projection results and the TSDF of the depth map are fused and input to the MVFS module to generate $R$ synthetic-view featuremaps. The $R$ feature maps are then fed into CVTr to capture more global information to obtain augmented-view feature mpas. Finally, the final dense semantic scene completion result is obtained through a classification head after upsampling by 3D CNN.}
%\label{fig:overview}
%\vspace{-0.1in}
%\end{figure*}

\begin{figure*}[t!]
\centering
\includegraphics[width=\linewidth]{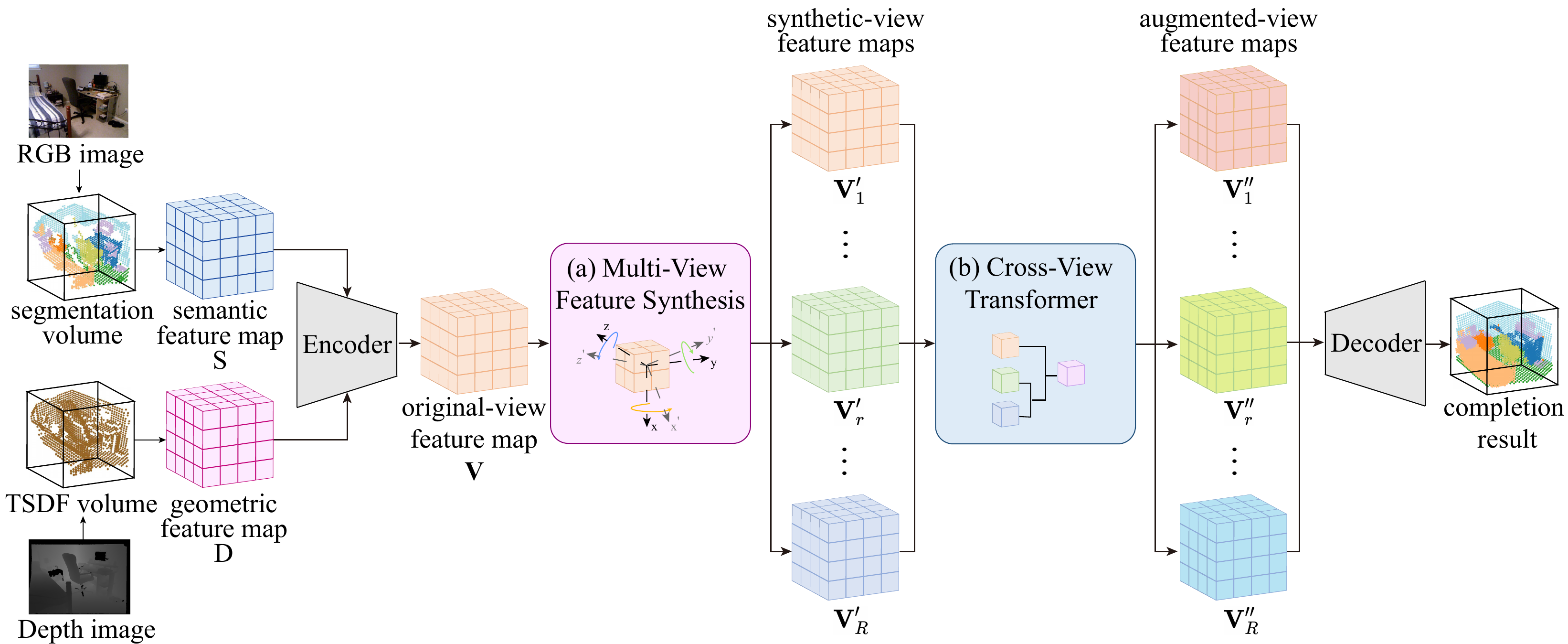}
\vspace{-0.1in}
\caption{The overall architecture of CVSformer, which consists of Multi-View Feature Synthesis (MVFS) and Cross-View Transformer (CVTr). Based on a single-view RGB-D image, CVSformer computes the semantic and geometric feature maps, which are input into the MVFS (a) to achieve multiple synthetic-view feature maps. (b) CVTr takes input as the synthetic-view feature maps, yielding augmented-view feature maps for the final completion task.}
\label{fig:overview}
\vspace{-0.2in}
\end{figure*}

In this paper, we propose \textbf{Cross-View Synthesis Transformer} (CVSformer), which consists of \textbf{Multi-View Feature Synthesis} (MVFS) and \textbf{Cross-View Transformer} (CVTr), as illustrated in Figure~\ref{fig:overview}. CVSformer employs the basic 3D CNN to learn the visual feature map from RGB-D image. The original-view feature map contains the feature of each voxel, whose correlation with the adjacent voxels is captured from the original view. Next, MVFS controls the rotation of the 3D convolutional kernel, synthesizing the change of view direction and providing new object relationships. Here, we remark that 3D kernel's view is an analogy with the camera's view. It has a conceptual view direction to determines the center voxel's spatial neighbors and their correlations. MVFS employs the kernels with different rotations to compute the synthetic-view feature maps, which capture the object relationships for each voxel from multiple views. Finally, all synthetic-view feature maps are input to CVTr, which conducts the cross-view fusion to establish the object relationships across multiple views. Based on the cross-view object relationships, CVTr enhances the synthetic-view feature maps, yielding the augmented-view feature maps for SSC.

We evaluate CVSformer on the public datasets for SSC, where we achieve the state-of-the-art results (i.e., 52.6 and 63.9 mIoUs on NYU~\cite{nyu} and NYUCAD~\cite{nyucad} datasets).

%We also comprehensively examine the effectiveness of all network components (i.e., MVFS and CVTr), which are essential for improving the completion results.

\section{Related work}

Below, we mainly survey the literature on semantic scene completion and multi-view information fusion, which highly correlate to our method in modeling the geometric and semantic relationships between 3D objects.

\subsection{Semantic Scene Completion}

SSCNet~\cite{sscnet} is a pioneering work that uses the deep network for semantic scene completion. It generates a complete 3D voxel representation of volumetric occupancy and semantic labels based on a single-view depth map. AICNet~\cite{aicnet} employs an anisotropic convolution module to balance performance and computational cost. 3D sketch~\cite{3dsketch} introduces CVAE in the network to guide sketch generation. SISNet~\cite{sisnet} lets the instance-completion and scene-completion promote mutually to obtain better completion results. CCPNet~\cite{ccpnet} has a cascaded architecture to achieve multi-scale 3D context and integrate local geometric details of the scene. SATNet~\cite{satnet} decomposes the semantic scene completion task into 2D semantic segmentation and 3D semantic scene completion, whose outputs are fused. ESSCNet~\cite{sgc} divides the input voxels into different groups and performs 3D sparse convolution separately. DDRNet~\cite{ddrnet} has a lightweight decomposition network to better fuse multi-scale features. FFNet~\cite{ffnet} obtains augmented features by modeling the frequency domain correlation between RGB and depth data. ForkNet~\cite{forknet} generates new samples to assist the training process.

Currently, semantic scene completion relies on the input of single-view images, which provide little information to establish the connection between the occluded objects and their neighbors. In contrast, we propose the multi-view feature synthesis by controlling the rotated kernels, which are used to extract the geometric and semantic representations for modeling the multi-view object relationships.

\subsection{Multi-View Information Fusion}

Many works on multi-view information fusion have combined the features learned from bird's eye view (BEV) and range view (RV) images. S3CNet~\cite{s3cnet} resorts to a set of bird's eye views to assist scene reconstruction. VISTA~\cite{vista} has dual cross-view spatial attention for fusing the multi-view image features. MVFuseNet~\cite{mvfusenet} utilizes the sequential multi-view fusion network to learn the image features from the RV and BEV views. 3DMV~\cite{3dmv} leverages a differentiable back-projection layer to incorporate the semantic features of RGB-D images. CVCNet~\cite{cvcnet} unifies BEV and RV, using a transformer to merge the two views. MVCNN~\cite{mvcnn} combines features of the same 3D shape from multiple perspectives. MV3D~\cite{mv3d} uses the region-based representation to fuse multi-scale features deeply.

The above methods either fuse BEV and RV of objects or fuse features of RGB-D sequences. Yet, these methods are less applicable in considering the object relationships across various views, which offer critical object context for semantic scene completion. Our framework uses kernels with different rotations to synthesize multi-view features based on a single-view image. Furthermore, we feed the multi-view features into the cross-view transformer, which communicates the multi-view features and yields the augmented information of cross-view object relationships for the completion task.

\section{Method Overview}

We illustrate the overall architecture of CVSformer in Figure~\ref{fig:overview}. At first, CVSformer takes the RGB image and the depth image as input. It projects the 2D semantic segmentation map of the RGB image into the voxelized segmentation volume~\cite{satnet}, where each voxel is associated with a category. Meanwhile, CVSformer computes the voxelized TSDF volume based on the depth map, where each voxel in the TSDF volume has a signed distance to the nearest object surface. We perform 3D convolution on the segmentation and TSDF volumes, computing the 3D semantic feature and geometric maps ${\bf S}, {\bf D} \in \mathbb{R}^{H \times W \times D \times C}$. Here, $H$, $W$, and $D$ represents the spatial resolutions along $x-$, $y-$, and $z-$axes\footnote{$x-$, $y-$, and $z-$axes are pre-defined by the ground-truth data.}. $C$ is the number of channels.

%Next, CVSformer adds the 3D semantic feature and geometric maps ${\bf S}$ and ${\bf D}$ and passes them into an encoder architecture with multiple 3D convolutional layers for computing the original-view feature map ${\bf V} \in \mathbb{R}^{H \times W \times D \times C}$. \textbf{Multi-View Feature Synthesis} (MVFS) takes ${\bf V}$ as input. As illustrated in Figure~\ref{fig:overview}(a), MVFS outputs a set of synthetic-view feature maps $\{{\bf V}'_r \in \mathbb{R}^{H \times W \times D \times C} ~|~ r=1,...,R\}$. Each synthetic-view feature map is computed by convoluting a set of 3D kernels, which are rotated by a controllable degree (see Figure~\ref{fig:MVFS}). \textcolor{red}{Note that the angles used in MVFS are based on the world coordinate system, where the positive $x-$axis is perpendicular to the screen pointing at the user, the positive $y-$axis is horizontal to the right side of the screen, and the positive $z-$axis is vertical up.}

Next, CVSformer adds the 3D semantic feature and geometric maps ${\bf S}$ and ${\bf D}$ and passes them into an encoder architecture with multiple 3D convolutional layers for computing the original-view feature map ${\bf V} \in \mathbb{R}^{H \times W \times D \times C}$. \textbf{Multi-View Feature Synthesis} (MVFS) takes ${\bf V}$ as input. As illustrated in Figure~\ref{fig:overview}(a), MVFS outputs a set of synthetic-view feature maps $\{{\bf V}'_r \in \mathbb{R}^{H \times W \times D \times C} ~|~ r=1,...,R\}$. Each synthetic-view feature map is computed by convoluting a set of 3D kernels, which are rotated by a controllable degree (see Figure~\ref{fig:MVFS}).

As illustrated in Figure~\ref{fig:overview}(b), all of the synthetic-view feature maps are fed into \textbf{Cross-View Transformer} (CVTr). CVTr employs cross-view fusion to let the synthetic-view feature maps mutually augment each other in an all-for-one fashion. Rather than straightforwardly fusing multiple synthetic-view feature maps, where the high-dimensional feature represents each voxel, CVTr computes separate view-tokens for the synthetic-view feature maps (see Figure~\ref{fig:cross_pipeline}). Each view token is low-dimensional. It represents the complete information of the corresponding synthetic-view feature map. CVTr resorts to the view-tokens to enhance all of the synthetic-view feature maps, yielding a set of augmented-view feature maps $\{{\bf V}''_r \in \mathbb{R}^{H \times W \times D \times C} ~|~ r=1,...,R\}$.

Eventually, we concatenate the augmented-view feature maps, feeding them into a 3D-convolutional decoder. The decoder outputs the voxelized volume, where each voxel has a semantic category as the completion result.

\section{Architecture of CVSformer}

In this section, we provide more details of the critical components of CVSformer (i.e., Multi-View Feature Synthesis and Cross-View Transformer).

\subsection{Multi-View Feature Synthesis}

MVFS controls the rotation degree of 3D convolutional kernels. It convolutes the rotated kernels with the original-view feature map, yielding the synthetic-view feature maps.

\begin{figure}[t!]
  \centering
  \includegraphics[width=\linewidth]{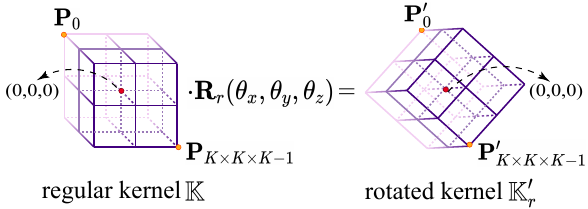}
  \vspace{-0.1in}
  \caption{The illustration of controlling the degree to rotate a 3D convolutional kernel. We fix the kernel center during rotation.}
  \label{fig:matrix_re}
  \vspace{-0.15in}
  \end{figure}

\vspace{0.1in}
\noindent{\bf Kernel Rotation~~}
We explain how to control the rotation of the 3D convolutional kernel in Figure~\ref{fig:matrix_re}. In the left of Figure~\ref{fig:matrix_re}, we denote a $K \times  K \times K$ 3D kernel as a set of 3D points $ \mathbb K = \{{\bf P}_{k}\in \mathbb{R}^3 ~|~ k  =  0, \cdots,K \times  K \times  K - 1 \}$. ${\bf P}_{k}$ contains $x-$, $y-$, and $z-$coordinates, which are computed as:
\begin{equation}\label{eq:matrix}
\left\{
\begin{aligned}
    \begin{array}{l}
    {\bf x}_{k}= \frac{2(k \bmod K^{2}\bmod K)-{K-1}}{2} \vspace{2ex}, \\
    {\bf y}_{k}= \frac{2(k \bmod K^{2}) -K({K-1})}{2K} \vspace{2ex}, \\
    {\bf z}_{k}= \frac{-2k  +K^{2}({K-1})}{{2}{K^{2}}} \vspace{2ex}, \\
    {\bf P}_{k}=  ({\bf x}_{k},{\bf y}_{k},{\bf z}_{k}),
\end{array}
\end{aligned}
\right.
\end{equation}
where ${\bf P}_0 = \left(-\frac{K-1}{2},-\frac{K-1}{2}, \frac{K-1}{2}\right)$, ${\bf P}_{\frac{K \times K \times K - 1}{2}} = \left(0,0,0\right)$, and ${\bf P}_{K \times K \times K - 1} = \left(\frac{K-1}{2},\frac{K-1}{2}, -\frac{K-1}{2}\right)$ represent the top-left, center, and bottom-right vertexes of the 3D kernel $\mathbb K$, respectively. And mod means the remainder operator with a left-to-right precedence.

\begin{figure*}[t!]
\centering
\includegraphics[width=\linewidth]{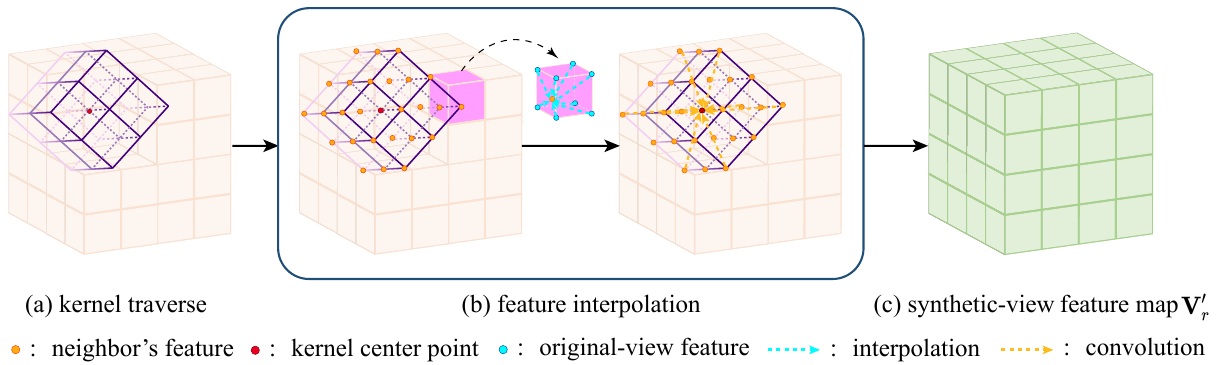}
\vspace{-0.2in}
\caption{The architecture of MVFS. It consists of two parts: Kernel Rotation and Interpolation. We calculate the synthetic-view feature maps by convoluting the different rotated kernels to the feature map of the original view.}
\vspace{-0.1in}
\label{fig:MVFS}
\end{figure*}

As illustrated in the right of Figure~\ref{fig:matrix_re}, we rotate the kernel $\mathbb K$ around $x-$, $y-$, and $z-$axes centered on ${\bf P}_{\frac{K \times K \times K - 1}{2}}$. For this purpose, we prepare a set of hybrid rotation matrices $\{{\bf R}_r\left(\theta_{x}, \theta_{{y}}, \theta_{{z}}\right) \in \mathbb{R}^{3 \times 3}~|~r=1,...,R\}$. We compute the $r^{th}$ rotation matrix ${\bf R}_r\left(\theta_{x}, \theta_{{y}}, \theta_{{z}}\right)$ as:
\begin{equation}\label{eq:matrix}
    \left\{
    \begin{aligned}
        \begin{array}{l}
        {\bf R}_r\left({\theta}_{{x}}\right)=\left[\begin{array}{ccc}
        -\cos \theta_{{x}} & \sin \theta_{{x}} & 0 \\
        -\sin \theta_{{x}} & -\cos \theta_{{x}} & 0 \\
        0 & 0 & 1
        \end{array}\right]\vspace{2ex}, \\
        {\bf R}_r\left(\theta_{{y}}\right)=\left[\begin{array}{ccc}
        \hspace{0.8em} \cos \theta_{y} & \hspace{0.5em}0 & \hspace{0.6em} \sin \theta_{y} \\
        0 & \hspace{0.5em}1 & 0 \\
        -\sin \theta_{y} & \hspace{0.5em}0 & \hspace{0.6em}\cos \theta_{y}
        \end{array}\right]\vspace{2ex}, \\
        {\bf R}_r\left(\theta_{{z}}\right)=\left[\begin{array}{ccc}
        \hspace{0.1em}1 & 0 & 0 \\
        \hspace{0.1em}0 & -\cos \theta_{z} & \sin \theta_{z} \\
        \hspace{0.1em}0 & -\sin \theta_{z} & -\cos \theta_{z}
        \end{array}\right]\vspace{2ex}, \\
        {\bf R}_r\left(\theta_{x}, \theta_{{y}}, \theta_{{z}}\right)={\bf R}_r\left(\theta_{x}\right) \cdot {\bf R}_r\left(\theta_{{y}}\right) \cdot {\bf R}_r\left(\theta_{{z}}\right),
    \end{array}
    \end{aligned}
    \right.
    \end{equation}
where $\theta_x$, $\theta_y$, $\theta_z$ represent the rotation degrees around the $x-$, $y-$, and $z-$axes, respectively. We use ${\bf R}_r\left(\theta_{x}, \theta_{{y}}, \theta_{{z}}\right)$ to rotate the vertexes in the kernel $ \mathbb K$ as:
\begin{equation}
\begin{aligned}
{\bf P}'_{r, k} = {\bf P}_{k} \cdot  {\bf R}_r\left(\theta_{x}, \theta_{y}, \theta_{z}\right),
\end{aligned}
\end{equation}
where we form a new kernel $ \mathbb{K}'_r = \{{\bf P}'_{r, k} \in \mathbb{R}^3 ~|~ k  =  0, \cdots,K \times  K \times  K - 1 \}$ with a rotated view.

\vspace{0.1in}
\noindent{\bf Feature Interpolation for Rotated Kernel~~}
We convolute the rotated 3D kernels with the original-view feature map ${\bf V} \in \mathbb{R}^{H \times W \times D \times C}$ to compute a set of synthetic-view feature maps $\{{\bf V}'_r \in \mathbb{R}^{H \times W \times D \times C} ~|~ r=1,...,R\}$. We provide the details of using the rotated kernel $ \mathbb{K}'_r$ to compute the synthetic-view feature map ${\bf V}'_r$ in Figure~\ref{fig:MVFS}. During the convolution, we traverse the center of the rotated kernel $ \mathbb{K}'_r$ along $x-$, $y-$, and $z-$axes of the original-view feature map ${\bf V}$ (see Figure~\ref{fig:MVFS}(a)).

Given the $i^{th}$ vertex ${\bf v}_i$ of ${\bf V}$ (see Figure~\ref{fig:MVFS}(b)), which overlaps with the center of the rotated kernel $ \mathbb{K}'_r$, we achieve $K \times K \times K$ neighbors of ${\bf v}_i$. We denote the neighbors as a set of 3D points $\{\widetilde{\bf v}_{i,k} \in \mathbb{R}^3 ~|~k=1,...,K \times  K \times  K - 1\}$, where we compute $\widetilde{\bf v}_{i,k}$ as:
\begin{equation}
\begin{aligned}
\widetilde{\bf v}_{i,k} = {\bf v}_i + {\bf P}'_{r, k}.
\end{aligned}
\end{equation}
Different from any vertex of the feature map ${\bf V}$, the feature located at the vertex $\widetilde{\bf v}_{i,k}$ may be unavailable. To enable a valid convolution on the set $\mathbb N\left({\bf v}_i\right)$, we resort to the feature interpolation to approximate the feature $\widetilde{\bf V}_{i,k} \in \mathbb{R}^C$ of the vertex $\widetilde{\bf v}_{i,k}$. The interpolation uses the vertexes of the unit cube, where the $\widetilde{\bf v}_{i,k}$ resides, to compute $\widetilde{\bf V}_{i,k}$ as:
\begin{equation}
\begin{aligned}
& \widetilde{\bf V}_{i,k} = \sum^{K \times  K \times  K - 1}_{k=0} (1 - | {\bf v}_j - \widetilde{\bf v}_{i,k} |) \cdot {\bf V}_j, \\
&~~~~~ s.t.~~({\bf v}_j - \widetilde{\bf v}_{i,k})^{\top}({\bf v}_j - \widetilde{\bf v}_{i,k}) \le 1.
\end{aligned}
\end{equation}
Here, ${\bf v}_j$ is the $j^{th}$ vertex of the feature map ${\bf V}$. ${\bf v}_j$ is associated with the feature ${\bf V}_j \in \mathbb{R}^C$. The feature interpolation yields a set of features $\mathbb N\left({\bf v}_i\right) = \{\widetilde{\bf V}_{i,k} \in \mathbb{R}^C~|~k=1,...,K \times  K \times  K - 1\}$ centered at the vertex ${\bf v}_i$. We convolute the rotated kernel $\mathbb{K}'_r$ with the feature set $\mathbb N\left({\bf v}_i\right)$ as:
\begin{equation}
\begin{aligned}
& {\bf V}'_{r,i} = \sum^{K \times  K \times  K - 1}_{k=0} {\bf w}'_{r, k} \cdot \widetilde{\bf V}_{i,k},
\end{aligned}
\end{equation}
where ${\bf w}'_{r, k}$ is the weight associated with the vertex ${\bf P}'_{r, k}$ of the rotated kernel $ \mathbb{K}'_r$. ${\bf V}'_{r,i} \in \mathbb{R}^C$ represents the feature of the vertex ${\bf v}_i$ of the synthetic-view feature map ${\bf V}'_r$.

\begin{figure*}[t!]
	\centering
	\includegraphics[width=\linewidth]{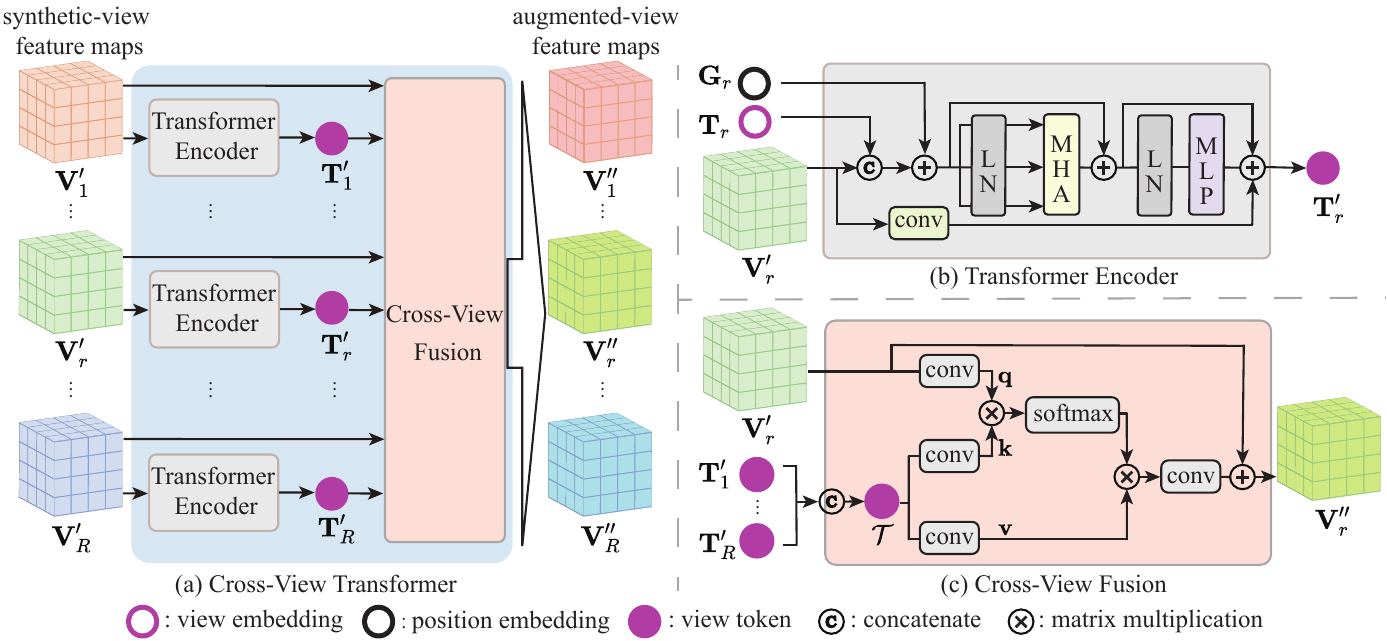}
	\vspace{-0.15in}
	\caption{(a) The overall architecture of CVTr. CVTr has several transformer encoders (b) for computing a low-dimensional view token. All view tokens are used by the cross-view fusion (c) for computing the augmented-view feature map.}
	\label{fig:cross_pipeline}
\vspace{-0.1in}
\end{figure*}

\subsection{Cross-View Transformer}

We illustrate the architecture of CVTr in Figure~\ref{fig:cross_pipeline}(a). We elaborate on the key modules of CVTr, i.e., transformer encoder and cross-view fusion in Figure~\ref{fig:cross_pipeline}(b--c).

\vspace{0.05in}
\noindent{\bf Transformer Encoder~~}
We illustrate a transformer encoder in Figure~\ref{fig:cross_pipeline}(b).
We input the synthetic-view feature map ${\bf V}'_r \in \mathbb{R}^{H \times W \times D \times C}$, the learnable view embedding ${\bf T}_r \in \mathbb{R}^{M \times C}$, and the learnable position embedding ${\bf G}_r \in \mathbb{R}^{(M+H \times W \times D) \times C}$ into the $r^{th}$ encoder. $M$ represents the resolution of the view embedding. The encoder outputs a low-dimensional view token ${\bf T}'_r \in \mathbb{R}^{M \times C}$ for representing the entire ${\bf V}'_r$ with a higher dimension.
% We flatten the resolution $H \times W \times D$ of the $(H \times W \times D) \times C$ feature map and concatenate it with the view token, forming a $(M+H \times W \times D) \times C$ feature map.

The encoder uses convolution to transform the synthetic-view feature map ${\bf V}'_r$, which is added to the view token ${\bf T}'_r$. This step is critical to the completion task based on the voxelized structure. It helps the view token ${\bf T}'_r$ with lower dimension to capture the underlying configuration of voxels.

The spatial dimension of the view embedding ${\bf T}_r$ is lower than the synthetic-view feature map ${\bf V}'_r$ (i.e., $M < H \times W \times D$). By compressing the spatial dimension of ${\bf T}_r$, we drive the ${\bf T}_r$ to focus on the global semantic meaning of ${\bf V}'_r$, which is injected into the view token ${\bf T}'_r$.

The position embedding ${\bf G}_r$ are jointly learned with the view embedding ${\bf T}_r$ and the synthetic-view feature map ${\bf V}'_r$. Thus, ${\bf G}_r$ can be regarded as a complementary structure, which stores richer geometric information than the view embedding ${\bf T}_r$. We also inject the geometric information of ${\bf G}_r$ into the view token ${\bf T}'_r$.

\vspace{0.05in}
\noindent{\bf Cross-View Fusion~~}
For all views, we use different transformer encoders to compute the view tokens $\{{\bf T}'_r \in \mathbb{R}^{M \times C} ~|~ r=1,...,R\}$. The cross-view fusion harnesses the view tokens to enhance the synthetic-view feature maps.

We provide more details of cross-view fusion in Figure~\ref{fig:cross_pipeline}(c). First, we concatenate all of the view tokens, forming an overall token $\mathcal{T} \in \mathbb{R}^{M \times R \times C}$ with hybrid information across different views. Next, we establish cross attention between the overall token $\mathcal{T}$ and the synthetic-view feature map ${\bf V}'_r$. With the cross-view information of $\mathcal{T}$, the cross attention enhances ${\bf V}'_r$ and computes the augmented-view feature map ${\bf V}''_r \in \mathbb{R}^{H \times W \times D \times C}$:
\begin{equation}
\left\{
\begin{aligned}
    \begin{array}{l}
    {\bf q} = conv({{\bf V}'_r}),~{\bf k} = conv({\mathcal{T}}),~{\bf v} = conv({\mathcal{T}}) \vspace{2ex}, \\
    {\bf A} = {\bf v}^{\top} \cdot softmax({\bf k} \cdot {\bf q}^{\top})  \vspace{2ex}, \\
    {\bf V}''_r = {\bf V}'_r + conv({\bf A}^{\top}).
\end{array}
\end{aligned}
\right.
\end{equation}
To simplify the notations above, we omit the subscripts of the intermediate variables ${\bf k}, {\bf v} \in \mathbb{R}^{M \times R \times C}$ and ${\bf q}, {\bf A} \in \mathbb{R}^{H \times W \times D \times C}$. CVTr produces a set of augmented-view feature maps $\{{\bf V}''_r \in \mathbb{R}^{H \times W \times D \times C}~|~r=1,...,R\}$, which are added together and passed to a 3D convolutional decoder for predicting the complete scene.

\section{Experiments}

\subsection{Implementation Details}

%We use PyTorch toolkit\footnote{https://pytorch.org/} to construct CVSformer with a NVIDIA 3090 GPUs. Specifically, we use DeepLabv3~\cite{deeplabv3} to pretrain the 2D semantic segmentation model for 1,000 epochs. Then we fix the pre-trained segmentation weights to train the CVSformer. We employ Stochastic Gradient Descent solver to optimize the parameters of CVSformer with a momentum of 0.9 and a weight decay of 0.0005. We set the initial learning rate to 0.05 and exploit a poly learning rate scheduler to dynamically adjust learning rate which is changed as $lr=init\_lr \times (1-\frac{epoch}{max\_epoch})^{0.9} $. The scene completion is supervised by voxel-wise cross-entropy loss. We train the network for $500$ epoches on the NYU and NYUCAD. And $16$ batch size per GPU. The spatial resolution of the projection result and TSDF is $60 \times 36\times 60$. The size of convolutional kernels of the cross-view fusion is $1 \times 1 \times 1$. In the MVFS module we use four rotation angles of $0^\circ$, $45^\circ$, $90^\circ$ and $135^\circ$, which are chosen based on our experience.

We use PyTorch\footnote{https://pytorch.org/} to construct CVSformer. Before training CVSformer, we pre-train DeepLabv3~\cite{deeplabv3} for 1,000 epochs to segment the RGB image. Below, we fix the optimized network parameters of DeepLabv3.

The voxel-wise cross-entropy loss supervises the scene completion task. We employ SGD solver to optimize CVSformer with a momentum of 0.9 and a weight decay of 0.0005. We set the initial learning rate to 0.05, which is adjusted by the poly scheduler. We train the network for $500$ epoches, where each min-batch contains $16$ samples.

%The spatial resolution of the segmentation and TSDF volumes are $60 \times 36\times 60$. The kernels with different rotations are $3 \times 3 \times 3$. In MVFS, we empirically choose the rotation degrees along $x-$, $y-$, and $z-$axes from the set $\{0^\circ, 45^\circ, 90^\circ, 135^\circ\}$. In CVTr, we set the spatial dimension $M=75$ of the view token as default.

The spatial resolution of the segmentation and TSDF volumes are $60 \times 36\times 60$, and the resolution of original-view feature map is $15 \times 9\times 15$. In MVFS, the rotated kernels are $3 \times 3 \times 3$. We empirically choose the rotation degrees along $x-$, $y-$, and $z-$axes from the set $\{0^\circ, 45^\circ, 90^\circ, 135^\circ\}$. To simplify the experiment, we empirically rotate the 3D kernel along $x-$axis. In CVTr, we set the spatial dimension $M\!=\!75$ of the view token as default.

\begin{figure*}[th!]
  \centering
  \begin{tabular}{@{\hspace{0mm}}c@{\hspace{1mm}}c@{\hspace{1mm}}c@{\hspace{1mm}}c}
  \includegraphics[width=0.245\linewidth]{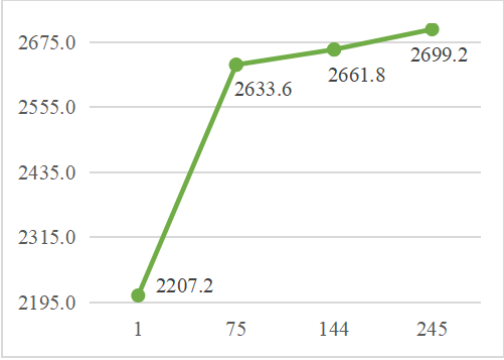} &
  \includegraphics[width=0.245\linewidth]{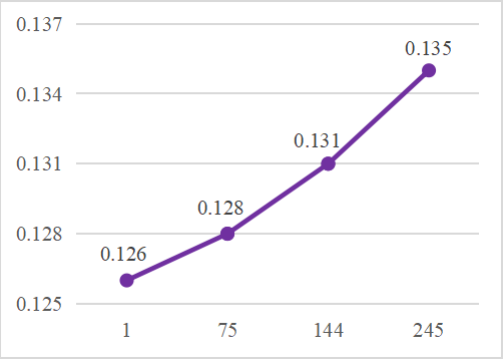} &
  \includegraphics[width=0.245\linewidth]{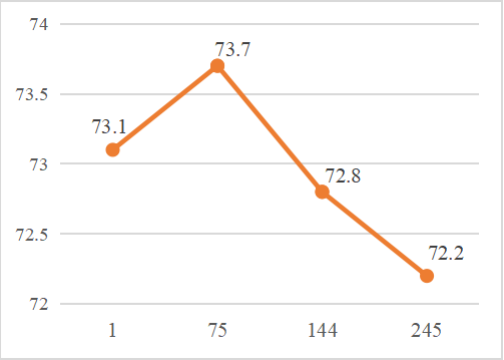} &
  \includegraphics[width=0.245\linewidth]{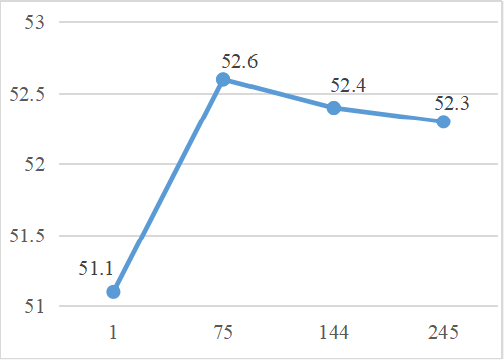} \\
  \small{(a) GPU Memory} & \small{(b) Testing Time} & \small{(c) IoU(\%)} & \small{(d) mIoU(\%)}
  \end{tabular}
  \vspace{0.01in}
  \caption{Sensitivities of GPU memory in MB (a), testing time per image in seconds (b), IoU (c), and mIoU (d) to the spatial resolution of view token. The performances are evaluated on NYU dataset.}
  \label{fig:curve_vt}
  \vspace{-0.2in}
  \end{figure*}
  
\subsection{Experimental Datasets}

\vspace{0.01in}
\noindent{\bf Datasets~~}
We compare CVSformer with state-of-the-art methods on the real NYU~\cite{nyu} and NYUCAD~\cite{nyucad} datasets. They both provide 1,449 pairs of RGB-D images, and there are $795$/$654$ pairs for training and test. We use the 3D annotation provided in~\cite{completing} for the network training.

\vspace{0.05in}
\noindent{\bf Evaluation Metrics~~}
We report the accuracies of scene completion (SC) and semantic scene completion (SSC) on different datasets. We use the recall, precision, and voxel-wise intersection-over-union (IoU) on the occluded voxels to measure the accuracy of SC. To calculate the accuracy of SSC, we report IoUs on different semantic categories, which are averaged to achieve mean IoU.

\subsection{Internal Study of CVSformer}

We use NYU~\cite{nyu} for examining the effectiveness of CVSformer and its key components (i.e., MVFS and CVTr).

\vspace{0.01in}
\noindent{\bf Spatial Resolution of View Tokens~~}
%The length of view token depends on the recognition level required by the specific task. A [$class$] token of length 1 in ViT~\cite{vit} is suitable for image-level tasks, such as image classification. However, we think it is not sufficient for dense voxel-wise recognition tasks, such as semantic scene completion. In order to study how long the view token has the ability to represent a single view of 3D scene, we choose the length of view token $M$ from the set $\{1, 75, 144, 245\}$, which corresponds to $\{1\times1\times1, 5\times3\times5, 6\times4\times6, 7\times5\times7\}$ voxel resolution respectively. We report the calculation overheads (i.e., GPU memory and testing time) in Figure~\ref{fig:curve_vt}(a-b), along with the completion performances (i.e., SC-IoU and SSC-mIoU) in Figure~\ref{fig:curve_vt}(c-d).
%
We change the spatial resolution of view tokens and examine the effect on the computational overheads (i.e., GPU memory and testing time in Figure~\ref{fig:curve_vt}(a--b)) and the performances (i.e., SC-IoU and SSC-mIoU in Figure~\ref{fig:curve_vt}(c--d)). We choose the spatial resolution from the set $\{1, 75, 144, 245\}$, which corresponds to $\{1\times1\times1, 5\times3\times5, 6\times4\times6, 7\times5\times7\}$ voxelized volumes, respectively. A higher resolution (e.g., $M=75$) allows view tokens to contain richer information of voxelized configuration, semantic meaning, and geometric relationship between objects. However, the view tokens with too high resolutions (e.g., $M=144$ and $245$) inevitably require more computation. They contain a too complex mixture of information. These view tokens are used by the cross-view fusion, where the complex information distracts the specific information of each view's feature map. Thus, they degrade the performances.

%With $M=1$, we only use the view token of length 1 to represent a whole view voxel, yielding 73.1$\%$ IoU and 51.1\% mIoU on the test set of NYU. This is because synthetic-view feature maps with the resolution of $15\times9\times15$ are relatively compact, such a small length of view token will result in a significant loss of scene information. By increasing the length of view token to 75, we successfully improve the performance by a large margin (up to 73.6$\%$ IoU and 52.6\% mIoU) on NYU, which also confirmed our idea. But the longer view token (e.g., $M=144, 245$) saturates the performances, at the cost of more computations. We exploit $M=75$ as default experiment setup and achieve stable improvement on NYUCAD dataset.

% \vspace{0.05in}
% \noindent{\bf Running time and model capacity~~}
% Compared to the state-of-the-art SISNet whose average running time and model capacity are 0.10s/scene and 314.6M, CVSformer requires 0.13s/scene and 430.5M.

\vspace{0.05in}
\noindent{\bf Different Strategies of Learning View Tokens~~}
%In Table~\ref{tab:CVTr_ablation_learning_method}, we experiment with different learning methods of the view token to evaluate their effectiveness. We first use 3D convolution and regular self attention method to learn the view token, in which mIoU only reached 51.5\% and 52.0\% respectively. With combining these two approaches, i.e., 3d convolution $+$ self attention, the completion performances are greatly improved (up to 73.7\% IoU and 52.6\% mIoU), as shown in the last row of Table~\ref{tab:CVTr_ablation_learning_method}. This is because combines 3D convolution with self attention, our learning method has the advantages of focusing on local relationships between voxels and capturing long-range dependencies of whole scene simultaneously.
%
In Table~\ref{tab:CVTr_ablation_learning_method}, we experiment with different strategies for learning view tokens. As reported in the first and second rows, we use the regular 3D convolution and self-attention to learn the view tokens, respectively, achieving (72.6\% IoU, 51.8\% mIoU) and (72.7\% IoU, 52.0\% mIoU). Note that self-attention slightly outperforms the regular 3D convolution because it is good at capturing long-range dependencies of the whole scene. Yet, it eliminates the configuration of the voxelized structure, which is respected by the regular 3D convolution. The importance of respecting the voxelized format is evidenced by the performance improvement up to (73.7\% IoU, 52.6\% mIoU) when we combine 3d convolution and self-attention for learning the view tokens.

\setlength{\tabcolsep}{8.0pt}
  \begin{table}[h!]
    \centering
    \begin{tabular}{c||cc}
    \hline
    \textbf{View Token}  & \textbf{IoU(\%)} & \textbf{mIoU(\%)}            \\ \hline \hline
    convolution       & 72.6          & 51.8 \\ \hline
    self-attention     & 72.7           & 52.0 \\ \hline
    convolution + self-attention & \textbf{73.7}        & \textbf{52.6} \\ \hline
    \end{tabular}
    \vspace{0.01in}
    \caption{Various strategies for learning view tokens. The performances are evaluated on NYU dataset.}
    \label{tab:CVTr_ablation_learning_method}
    \vspace{-0.02in}
  \end{table}

%\vspace{0.05in}
%\noindent{\bf Analysis of synthetic-view feature map quantities~~}
%Considering that the different number of synthetic-view feature maps generated by the MVFS module can lead to different object relationships for each voxel captured thus affecting the final semantic completion performance. Therefore, in Table~\ref{tab:branch_num}, we compare the impact on IoU and mIoU when this factor is varied. According to the data in the table, the best performance is obtained when MVFS generates four synthetic-view feature maps. As to why the performance is better when generating two new synthetic-view feature map than when generating three, we think it is because generating one more synthetic-view feature map will make the fusion with CVTr more challenging later, and the generated feature map has not provided enough new information, so the two sides do not reach a balance and thus the performance decreases. However, the performance of the second, third and fourth rows is higher than that of the first baseline. Therefore, we believe that MVFS performs best when it generates four new synthetic-view feature maps.

\vspace{0.05in}
\noindent{\bf Sensitivity to View Number in MVFS~~}
We experiment with changing the number of views, where each view is associated with a rotation degree in the set $\{0^{\circ},45^{\circ},90^{\circ},135^{\circ}\}$. Different views yield the corresponding synthetic-view feature maps. We report the performances in Table~\ref{tab:branch_num}. We find that more views saturate IoU sensitive to the object occupancies in the voxels. Comparably, multiple synthetic-view feature maps are used by CVTr to capture cross-view object relationships for comprehensively representing the semantic correlation between objects. Compared to the single view (i.e., $0^{\circ}$), more views improve mIoU up to 52.6\%, which requires differentiating the semantic object categories. More detailed discussion of MVFS can be found in the supplementary material.

\setlength{\tabcolsep}{13pt}
  \begin{table}[t!]
    \centering
    \begin{tabular}{c||cc}
    \hline
    \textbf{Rotation Degrees}  & \textbf{IoU(\%)} & \textbf{mIoU(\%)}            \\ \hline \hline
    $0^{\circ}$       & 72.9          & 51.3 \\ \hline
    $\{0^{\circ},45^{\circ}\}$     & \textbf{73.8}           & 52.1 \\ \hline
    $\{0^{\circ},45^{\circ},90^{\circ}\}$ & 73.5        & 52.0 \\ \hline
    $\{0^{\circ},45^{\circ},90^{\circ},135^{\circ}\}$ & 73.7      & \textbf{52.6} \\ \hline
    \end{tabular}
    \vspace{0.01in}
    \caption{Various number of views in MVFS. The performances are evaluated on NYU dataset.}
    \vspace{-0.02in}
    \label{tab:branch_num}
  \end{table}

%\vspace{0.05in}
%\noindent{\bf Different fusion schemes of CVTr~~}
%To evaluate the benefits of our cross-view fusion module, we compare with the alternatives in Table~\ref{tab:CVTr_ablation_fusion_scheme}.
%First, we perform the self attention block directly on all concatenated synthetic-view feature maps, i.e., all fusion scheme. Further, we try to use the concatenated synthetic-view feature maps to enhance the synthetic-view features individually by cross attention, where the single synthetic-view feature is query and the concatenated synthetic-view feature maps is the key and value. However, these schemes do not take into account the information differences in different views, leading to information disorder. Meanwhile, the computational cost of the two methods above grows quadratically with the size of query and value (i.e., $O(N)=N \times N$).

%With using the view tokens obtained from the transformer encoder to enhance the synthetic-view features individually, the completion performances are steadily improved, up to 73.7\% IoU and 52.6 mIoU respectively. The view tokens for all views contain information about the same voxelized configuration. Therefore, our method reduces the information difference for better cross-view fusion. And the calculation complexity is reduced to linear (i.e., $O(N)=N\times M,M\ll N$).

\vspace{0.05in}
\noindent{\bf Different Fusion Schemes in CVTr~~}
In Table~\ref{tab:CVTr_ablation_fusion_scheme}, we compare different alternatives of cross-view fusion. We concatenate all synthetic-view feature maps fed into self-attention for computing an augmented-view feature map (see ``all fusion"). This scheme yields the performances of (73.0\% IoU, 52.3\% mIoU).

\setlength{\tabcolsep}{7.0pt}
  \begin{table}[]
    \centering
    \begin{tabular}{c||cc}
    \hline
    \textbf{Fusion Scheme}  & \textbf{IoU(\%)} & \textbf{mIoU(\%)}            \\ \hline \hline
    all fusion     & 73.0           & 52.3 \\ \hline
    all-for-one w.r.t. feature maps       & 72.9          & 51.6 \\ \hline
    all-for-one w.r.t. tokens       & \textbf{73.7}          & \textbf{52.6} \\ \hline
    \end{tabular}
    \vspace{0.02in}
    \caption{Various fusion schemes in CVTr. The performances are evaluated on NYU dataset.}
    \label{tab:CVTr_ablation_fusion_scheme}
    \vspace{-0.1in}
    \end{table}

    \begin{figure*}[t!]
\centering
\includegraphics[width=\linewidth]{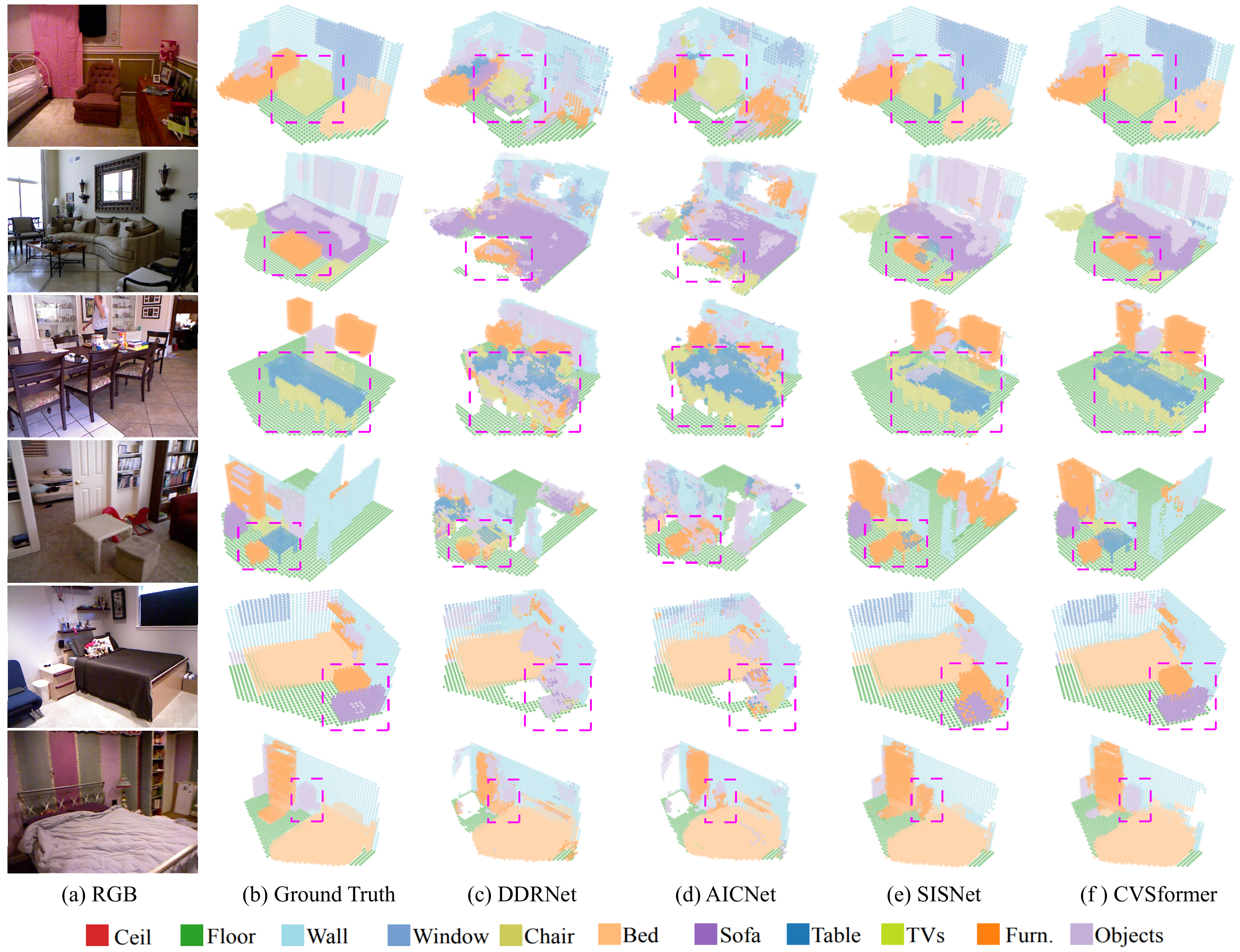}
\vspace{-0.19in}
\caption{Completion results of different methods. The examples are taken from the test set of NYU.}
\label{fig:visual_result}
\end{figure*}

In the second scheme (see ``all-for-one w.r.t. feature maps"), we concatenate all synthetic-view feature maps as the key and value of cross attention, which enhances each synthetic-view feature map. This scheme yields performances (72.9\% IoU, 51.6\% mIoU) worse than the first scheme. It evidences that high-dimensional synthetic-view feature maps contain redundant information for distracting the information fusion between multiple views.

Our cross-view fusion (see ``all-for-one w.r.t. tokens") employs the concatenated view tokens learned by the transformer encoders as the key and value for feature enhancement. The cross-view fusion refines the semantic and geometric information contained in the relatively low-dimensional view tokens, outperforming other alternatives.

  \setlength{\tabcolsep}{13.5pt}
  \begin{table}
    \begin{tabular}{c|c||cc}
    \hline
    \textbf{MVFS}     & \textbf{CVTr}     & \textbf{IoU(\%)}  & \textbf{mIoU(\%)}            \\ \hline \hline
                      &                   &    72.9           & 51.3 \\ \hline
    \checkmark        &                   & 73.2              & 51.8  \\ \hline
    \checkmark        & \checkmark        & \textbf{73.7}     & \textbf{52.6} \\ \hline
    \end{tabular}
    \vspace{0.01in}
    \caption{Ablation study on MVFS and CVTr. The performances are evaluated on NYU dataset.}
    \label{tab:component_result}
    \vspace{-0.1in}
    \end{table}

\setlength{\tabcolsep}{3.7pt}
\renewcommand{\arraystretch}{1.1}
\begin{table*}
\centering
\begin{threeparttable}
\begin{tabular}{c||ccc|ccccccccccc|c}
\hline
 \multirow{2}{*}{\textbf{Method}}   &  \multicolumn{3}{c|}{\textbf{SC}}  & \multicolumn{12}{c}{\textbf{SSC}}\\
 \cline{2-16}
& \multicolumn{1}{c}{\textbf{prec.}} & \multicolumn{1}{c}{\textbf{recall}} & \multicolumn{1}{c|}{\textbf{IoU}}& \multicolumn{1}{c}{\textbf{ceil.}}& \multicolumn{1}{c}{\textbf{floor}}& \multicolumn{1}{c}{\textbf{wall}}& \multicolumn{1}{c}{\textbf{win.}}& \multicolumn{1}{c}{\textbf{chair}}& \multicolumn{1}{c}{\textbf{bed}}& \multicolumn{1}{c}{\textbf{sofa}}& \multicolumn{1}{c}{\textbf{table}}& \multicolumn{1}{c}{\textbf{tvs}}& \multicolumn{1}{c}{\textbf{furn.}} & \multicolumn{1}{c|}{\textbf{objs.}} & \multicolumn{1}{c}{\textbf{avg.}}
 \\ \hline\hline
      SISNet(instance)~\cite{sisnet}  &92.1 &83.8 &78.2 &54.7 &93.8 &53.2 &41.9 &43.6 &66.2 &61.4 &38.1 &29.8 &53.9 &40.3 &52.4\\ \hline\hline
      SSCNet~\cite{sscnet}  &57.0 &\textbf{94.5} &55.1 &15.1 &94.7 &24.4 &0.0  &12.6 &32.1 &35.0 &13.0 &7.8  &27.1 &10.1 &24.7\\
      DDRNet~\cite{ddrnet}   &71.5 &80.8 &61.0 &21.1 &92.2 &33.5 &6.8 &14.8 &48.3 &42.3 &13.2 &13.9 &35.3 &13.2 &30.4\\
      AICNet~\cite{aicnet}   &62.4 &91.8 &59.2 &23.2 &90.8 &32.3 &14.8 &18.2 &51.1 &44.8 &15.2 &22.4 &38.3 &15.7 &33.3\\
      % TS3D~\cite{two}  &- &- &60.0 &9.7 &93.4 &25.5 &21.0 &17.4 &55.9 &49.2 &17.0 &27.5 &39.4 &19.3 &34.1\\
      SATNet~\cite{satnet}  &67.3 &85.8 &60.6 &17.3 &92.1 &28.0 &16.6 &19.3 &57.5 &53.8 &17.2 &18.5 &38.4 &18.9 &34.4\\
      % ForkNet~\cite{forknet}   &- &- &63.4 &36.2 &93.8 &29.2 &18.9 &17.7 &61.6 &52.9 &23.3 &19.5 &45.4 &20.0 &37.1\\
      CCPNet~\cite{ccpnet}   &74.2 &90.8 &63.5 &23.5 &\textbf{96.3} &35.7 &20.2 &25.8 &61.4 &56.1 &18.1 &28.1 &37.8 &20.1 &38.5\\
      Sketch~\cite{3dsketch}   &85.0 &81.6 &71.3 &43.1 &93.6 &40.5 &24.3 &30.0 &57.1 &49.3 &29.2 &14.3 &42.5 &28.6 &41.1\\

      SemanticFu~\cite{fu2022semantic}  &86.6 &82.4 &73.1 &45.4 &92.3 &41.1 &25.6 &32.6 &58.3 &49.8 &30.5 &17.1 &44.1 &33.9 &42.8\\

      FFNet~\cite{ffnet}  &89.3 &78.5 &71.8 &44.0 &93.7 &41.5 &29.3 &36.2 &59.0 &51.1 &28.9 &26.5 &45.0 &32.6 &44.4\\

      SISNet(voxel)~\cite{sisnet} &{88.7} &77.7 &70.8 &\textbf{46.8} &93.4 &42.0 &32.4 &36.0 &61.1 &55.8 &28.2 &27.6 &45.7 &32.9 &45.6\\

      PCANet~\cite{pcanet} &\textbf{89.5} &87.5 &\textbf{78.9} &44.3 &94.5 &\textbf{50.1} &30.7 &41.8 &\textbf{68.5} &56.4 &32.6 &29.9 &53.6 &\textbf{35.4} &48.9\\

  \textbf{CVSformer}   &87.7 &82.1 &{73.7} &46.3 &94.3&{43.5} & \textbf{42.8}& \textbf{46.2}& {67.7}& \textbf{66.0}& \textbf{39.2}& \textbf{43.2}& \textbf{53.9}&{35.0}  &\textbf{52.6}\\
\hline
\end{tabular}
\end{threeparttable}
\vspace{-0.01in}
\caption{We compare CVSformer with state-of-the-art methods on the test set of NYU.}
\label{tab:final_result_nyu}
\end{table*}

\setlength{\tabcolsep}{3.7pt}
\renewcommand{\arraystretch}{1.1}
\begin{table*}
\centering
\begin{threeparttable}
\begin{tabular}{c||ccc|ccccccccccc|c}
\hline
 \multirow{2}{*}{\textbf{Method}}    &  \multicolumn{3}{c|}{\textbf{SC}}  & \multicolumn{12}{c}{\textbf{SSC}}\\
 \cline{2-16}
& \multicolumn{1}{c}{\textbf{prec.}} & \multicolumn{1}{c}{\textbf{recall}} & \multicolumn{1}{c|}{\textbf{IoU}}& \multicolumn{1}{c}{\textbf{ceil.}}& \multicolumn{1}{c}{\textbf{floor}}& \multicolumn{1}{c}{\textbf{wall}}& \multicolumn{1}{c}{\textbf{win.}}& \multicolumn{1}{c}{\textbf{chair}}& \multicolumn{1}{c}{\textbf{bed}}& \multicolumn{1}{c}{\textbf{sofa}}& \multicolumn{1}{c}{\textbf{table}}& \multicolumn{1}{c}{\textbf{tvs}}& \multicolumn{1}{c}{\textbf{furn.}} & \multicolumn{1}{c|}{\textbf{objs.}} & \multicolumn{1}{c}{\textbf{avg.}}
 \\ \hline\hline
      SISNet(instance)~\cite{sisnet}  &94.1 &91.2 &86.3 &63.4 &94.4 &67.2 &52.4 &59.2 &77.9 &71.1 &51.8 &46.2 &65.8 &48.8 &63.5\\ \hline\hline
      SSCNet~\cite{sscnet}   &75.4 &\textbf{96.3} &73.2 &32.5 &92.6 &40.2 &8.9 &33.9 &57.0 &59.5 &28.3 &8.1 &44.8 &25.1 &40.0 \\
      DDRNet~\cite{ddrnet}   &88.7 &88.5 &79.4 &54.1 &91.5 &56.4 &14.9 &37.0 &55.7 &51.0 &28.8 &9.2 &44.1 &27.8 &42.8\\
      AICNet~\cite{aicnet}   &88.2 &90.3 &80.5 &53.0 &91.2 &57.2 &20.2 &44.6 &58.4 &56.2 &36.2 &9.7 &47.1 &30.4 &45.8\\
      % TS3D~\cite{two}  &- &- &76.1 &25.9 &93.8 &48.9 &33.4 &31.2 &66.1 &56.4 &31.6 &38.5 &51.4 &30.8 &46.2\\
      CCPNet~\cite{ccpnet}   &91.3 &92.6 &82.4 &56.2 &{94.6} &58.7 &35.1 &44.8 &68.6 &65.3 &37.6 &35.5 &53.1 &35.2 &53.2\\
      Sketch~\cite{3dsketch}   &90.6 &92.2 &84.2 &59.7 &94.3 &{64.3} &32.6 &51.7 &72.0 &68.7 &45.9 &19.0 &60.5 &38.5 &55.2\\

      SemanticFu~\cite{fu2022semantic} &88.7 &92.5 &84.8 &54.5 &94.8 &63.3 &29.3 &50.9 &73.6 &{70.9} &56.4 &31.7 &61.3 &42.0 &57.2\\

      FFNet~\cite{ffnet} &\textbf{94.8} &90.3 &85.5 &62.7 &\textbf{94.9} &\textbf{67.9} &35.2 &52.0 &74.8 &69.9 &47.9 &27.9 &62.7 &35.1 &57.4\\

      SISNet(voxel)~\cite{sisnet} &92.0 &89.3 &82.8 &62.0 &94.1 &63.3 &43.5 &50.8 &73.3 &63.5 &42.2 &40.6 &58.2 &39.7 &57.4\\

      PCANet~\cite{pcanet} &92.1 &91.8 &84.3 &54.8 &93.1 &62.8 &44.3 &52.3 &75.6 &70.2 &46.9 &44.8 &{65.3} &\textbf{45.8} &59.6\\

  \textbf{CVSformer}   &{94.0} &{91.0} &\textbf{86.0} &\textbf{65.6} &{94.2} &{60.6} & \textbf{54.7}& \textbf{60.4}& \textbf{81.8}& \textbf{71.3}& \textbf{49.8}& \textbf{55.5}& \textbf{65.5}&{43.5}  &\textbf{63.9}\\
\hline
\end{tabular}
\end{threeparttable}
\vspace{-0.01in}
\caption{We compare CVSformer with state-of-the-art methods on the test set of NYUCAD.}
\label{tab:final_result_nyucad}
\vspace{-0.1in}
\end{table*}
\vspace{-0.01in}

% IOU:
% [0.8242626878656254, 0.6559190528405069, 0.9417579785308028, 0.6061168923443716, 0.5467054796046427, 0.6037523382888649, 0.818102359117346, 0.7129529555365564, 0.49763145428706773, 0.5551824148984887, 0.6552167556126236, 0.434583085843235]

% meanIOU: 0.638902

% prec.: 0.839338

% --*-- Scene Completion --*--

% IOU: 0.859556

% prec.: 0.940135

% recall: 0.909327

\noindent{\bf Analysis of Network Components~~}
In Table~\ref{tab:component_result}, we examine the effectiveness of the critical components (i.e., MVFS and CVTr), by removing one or more components from the completion network. In the first row, we use the baseline network without decoration, which employs the primary 3D convolution to learn the original-view feature map from RGB-D image for semantic scene completion. This method produces (72.9\% IoU, 51.3\% mIoU).

We add MVFS to the network (see the second row) for computing multiple synthetic-view feature maps, which are concatenated for regressing the final result. MVFS helps to improve performance (73.2\% IoU, 51.8\% mIoU), demonstrating the necessity of multi-view information.

Next, we combine MVFS and CVTr (see the third row) to form the entire model. Here, we fuse the multi-view information to compute the augmented-view feature maps, further improving performance (73.7 IoU\%, 52.6\% mIoU). This result demonstrates that the trivial concatenation of synthetic-view feature maps in the second alternative is less effective in modeling the cross-view object relationship.

\vspace{-0.07in}
\subsection{Comparisons with State-of-the-Art Methods}
\vspace{-0.07in}
%We compare CVSformer with the recent state-of-the-art methods on the public real and synthetic datasets, i.e., NYU~\cite{nyu} and NYUCAD~\cite{nyucad}.

Below, we compare CVSformer with state-of-the-art methods on NYU~\cite{nyu} and NYUCAD~\cite{nyucad} datasets in Tables~\ref{tab:final_result_nyu} and~\ref{tab:final_result_nyucad}.

We compare CVSformer with the completion methods that rely on voxel-wise annotations for the network training. Here, CVSformer outperforms the current methods (e.g., Sketch~\cite{3dsketch} and SISNet(voxel)~\cite{sisnet}) by a remarkable margin. In Figure~\ref{fig:visual_result}, CVSformer achieves the visualized results better than the competitive methods. More results please refer to supplementary materials.

The latest SISNet(instance)~\cite{sisnet} takes advantage of the instance-wise annotations, which provide more substantial supervision than the voxel-wise annotations for the network training. It achieves remarkable performances, i.e., (78.2\% IoU, 52.4\% mIoU) on NYU and (86.3\% IoU, 63.5\% mIoU) on NYUCAD. CVSformer still outperforms SISNet(instance)~\cite{sisnet} by only requiring voxel-wise annotations on NYU and NYUCAD.

We also evaluate the computation overhead of CVSformer with other methods, in terms of GPU memory, running time, and model capacity. Compared to AICNet (3413M, 0.23s/scene, 0.72M), DDRNet (3007M, 0.14s/scene, 0.20M), and SISNet (3750M, 0.18s/scene, 0.57M), CVSformer (2633M, 0.13s/scene, 0.41M) requires reasonable computation.

\vspace{-0.015in}
\section{Conclusion}
\label{sec:conclusion}
\vspace{-0.02in}
%The latest progress in semantic scene completion benefits from deep neural networks to learn the geometric and semantic representations of 3D objects. In this paper, we have proposed a deep neural networks, CVSformer, to learn cross-view object relationships for semantic scene completion.
%CVSformer consists of Multi-View Feature Synthesis and Cross-View Transformer. In contrast to networks with single-view input, CVSformer controls kernels with different rotations to provide new object relationships. We employ a novel fusion scheme across multiple views to enhance the synthetic-view features. Finally, the obtained cross-view object relationships provide the relevant information for suggesting the existence of the occluded objects. CVSformer achieves state-of-the-art performances on the public datasets.

The latest progress in semantic scene completion benefits from deep neural networks to learn the geometric and semantic features of 3D objects. In this paper, we propose CVSformer that understands cross-view object relationships for semantic scene completion. CVSformer controls kernels with different rotations to learn multi-view object relationships. Furthermore, we utilize a cross-view fusion to exchange information across different views, thus capturing the cross-view object relationships. CVSformer achieves state-of-the-art performances on public datasets. In the future, we plan to explore how to use CVSformer for other 3D object recognition tasks.

%\section*{Acknowledgments}
%We thank the anonymous reviewers for their constructive comments.

{\small
\bibliographystyle{ieee_fullname}
\bibliography{CVSformer}
}

\end{document}